\documentclass[10pt,twocolumn,letterpaper]{article}

\usepackage{iccv}
\usepackage{times}
\usepackage{epsfig}
\usepackage{graphicx}
\usepackage{amsmath}
\usepackage{amssymb}
\usepackage{booktabs}
\usepackage{multirow}
\usepackage{color}
\usepackage[cmyk]{xcolor}

\usepackage[pagebackref=true,breaklinks=true,letterpaper=true,colorlinks,bookmarks=false]{hyperref}

\iccvfinalcopy 

\def\iccvPaperID{10661} 

\ificcvfinal\fi

\begin{document}

\title{Distilling Coarse-to-Fine Semantic Matching Knowledge

for Weakly Supervised 3D Visual Grounding}

\author{Zehan Wang\thanks{Equal contribution} \hspace{1em} Haifeng Huang\footnotemark[1] \hspace{1em} Yang Zhao \hspace{1em} Linjun Li \hspace{1em} Xize Cheng \\ Yichen Zhu \hspace{1em} Aoxiong Yin \hspace{1em} Zhou Zhao\thanks{Corresponding author}\\
Zhejiang University\\
{\tt\small \{wangzehan01, huanghaifeng, zhaozhou\}@zju.edu.cn}
}

\maketitle

\begin{abstract}
3D visual grounding involves finding a target object in a 3D scene that corresponds to a given sentence query. Although many approaches have been proposed and achieved impressive performance, they all require dense object-sentence pair annotations in 3D point clouds, which are both time-consuming and expensive. To address the problem that fine-grained annotated data is difficult to obtain, we propose to leverage weakly supervised annotations to learn the 3D visual grounding model, i.e., only coarse scene-sentence correspondences are used to learn object-sentence links. To accomplish this, we design a novel semantic matching model that analyzes the semantic similarity between object proposals and sentences in a coarse-to-fine manner. Specifically, we first extract object proposals and coarsely select the top-K candidates based on feature and class similarity matrices. Next, we reconstruct the masked keywords of the sentence using each candidate one by one, and the reconstructed accuracy finely reflects the semantic similarity of each candidate to the query. Additionally, we distill the coarse-to-fine semantic matching knowledge into a typical two-stage 3D visual grounding model, which reduces inference costs and improves performance by taking full advantage of the well-studied structure of the existing architectures. We conduct extensive experiments on ScanRefer, Nr3D, and Sr3D, which demonstrate the effectiveness of our proposed method.
\end{abstract}

\section{Introduction}
\label{sec:intro}
3D Visual grounding (3DVG) refers to the process of localizing an object in a scene based on a natural language sentence. The 3DVG task has recently gained attention due to its numerous applications. Despite the significant progress made in this area \cite{cai20223djcg, chen2020scanrefer, yuan2021instancerefer, zhao20213dvg, huang2022multi, yang2021sat}, all these approaches require bounding box annotations for each sentence query, which are laborious and expensive to obtain. For example, it takes an average of 22.3 minutes to annotate a scene in the ScanNet-v2 dataset \cite{dai2017scannet}. Thus, we focus on weakly supervised training for 3DVG, which only requires scene-sentence pairs for training. This problem is meaningful and realistic since obtaining scene-level labels is much easier and can be scaled effectively.

\begin{figure}[t]
\centering
\includegraphics[width=\linewidth]{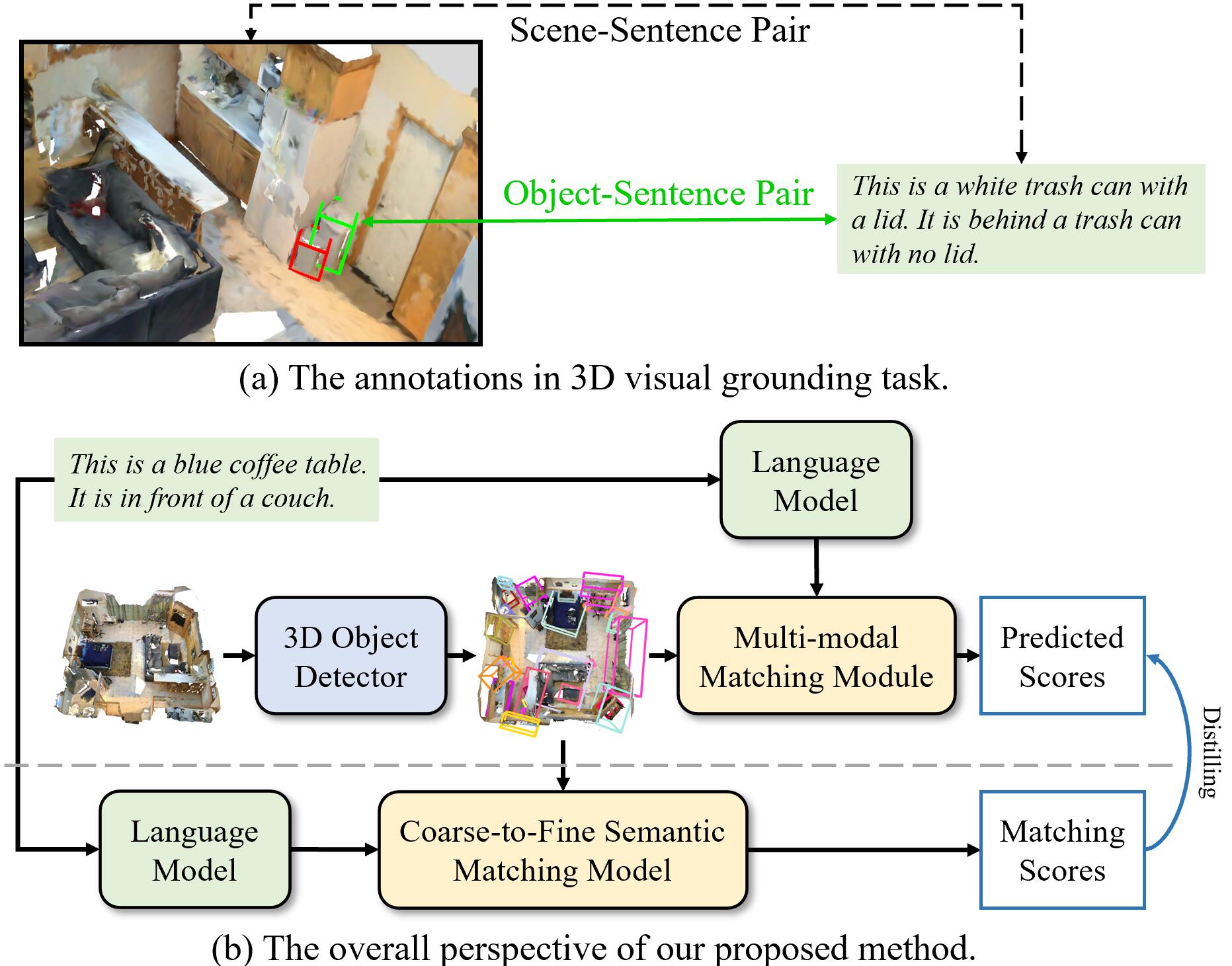}
\caption{(a). 3D visual grounding aims to find the object-sentence pair from the whole scene. The fully supervised setting requires all the dense ground-truth object-sentence labels for training, while the weakly supervised method only needs the coarse scene-sentence annotations. (b). Coarse-to-Fine Semantic Matching Model (bottom) analyzes the matching score of each proposal to the sentence, and the semantic matching knowledge is distilled to the two-stage 3DVG architecture (upper).}
\label{fig:illustration}
\end{figure}

However, weakly supervised 3DVG poses two challenges. Firstly, a 3D point cloud can contain numerous objects of various categories, and a sentence query may contain multiple objects besides the target object to aid in localization. Without knowledge of the ground-truth object-sentence pair, it is difficult to learn to link the sentence to its corresponding object from the enormous number of possible object-sentence pairs. Secondly, the 3DVG task often involves multiple interfering objects in the scene with the same class as the target object, and the target object must be distinguished based on its object attributes and the relations between objects described in the given sentence. As illustrated in Figure \ref{fig:illustration} (a), there are two trash cans in the scene, and the described target object can only be identified by fully comprehending the language description.

To address both challenges simultaneously, we propose a coarse-to-fine semantic matching model to measure the similarity between object proposals and sentences. Specifically, our model generates object-sentence matching scores from scene-sentence annotation, guided by coarse-to-fine semantic similarity analysis. Firstly, we calculate the object category similarity and feature similarity between all the proposals and the sentence. Combining these two similarities, we roughly select $K$ proposals with the highest similarity to the sentence, which can effectively filter out the proposals that do not belong to the target category. Secondly, we utilize NLTK~\cite{bird2009natural} to conduct part-of-speech tagging on the sentences and randomly mask the more meaningful nouns and adjectives words. The selected candidates would be used to reconstruct the masked keywords of the sentence, which can help the model fully and deeply understand the whole sentence. Since the target object and the sentence query are semantically consistent, the more the candidate and the target object overlap, the more accurate its predicted keywords will be. The object-sentence matching score of each candidate can be measured by its reconstruction loss. Eventually, in order to reduce inference time and make full use of the structure of existing 3DVG models, we utilize knowledge distillation~\cite{hinton2015distilling} to migrate the knowledge of the coarse-to-fine semantic matching model to a typical two-stage 3DVG model, where the distilled pseudo labels are generated by the object-sentence matching scores.


In summary, the key contribution is four-fold:
\begin{itemize}
    \item[$\bullet$] To the best of our knowledge, this paper is the first work to address weakly supervised 3DVG, which eliminates the need for expensive and time-consuming dense object-sentence annotations and instead requires only scene-sentence level labels.
    \item[$\bullet$] We approach weakly supervised 3DVG as a coarse-to-fine semantic matching problem and propose a coarse-to-fine semantic matching model to analyze the similarity between each proposal and the sentence.
    \item[$\bullet$] We distill the knowledge of the coarse-to-fine semantic matching model into a two-stage 3DVG model, which fully leverages the well-studied network structure design, leading to improved performance and reduced inference costs.
    \item[$\bullet$] Experiments conducted on three wide-used datasets ScanRefer~\cite{chen2020scanrefer}, Nr3D~\cite{achlioptas2020referit3d} and Sr3D~\cite{achlioptas2020referit3d} demonstrate the effectiveness of our method. 
\end{itemize}

\section{Related Work}
\label{sec:related_work}

\noindent \textbf{Supervised 3D Visual Grounding.}
Grounding a sentence query in a 3D point cloud is a fundamental problem in vision-language tasks, with wide-ranging applications in fields like automatic robotics~\cite{xia2018gibson, wang2019reinforced, mittal2020attngrounder, feng2021cityflow} and AR/VR/metaverse~\cite{mystakidis2022metaverse, dionisio20133d}. The ScanRefer~\cite{chen2020scanrefer} and Referit3D~\cite{achlioptas2020referit3d} datasets annotate dense object-sentence links on the widely-used 3D point cloud dataset ScanNet~\cite{dai2017scannet}.

Most recent 3D visual grounding methods~\cite{cai20223djcg, yuan2021instancerefer, zhao20213dvg, huang2022multi, huang2021text, yang2021sat} follow a two-stage pipeline. In the first stage, pre-trained 3D object detectors~\cite{qi2019deep, liu2021group} generate 3D object proposals. The second stage involves matching the selected object proposals with the sentence query. Existing two-stage methods improve performance by exploring the object attributes and relations between proposals in the second stage. For example, 3DVG-Transformer~\cite{zhao20213dvg} uses a coordinate-guided contextual aggregation module to capture relations between proposals and a multiplex attention module to distinguish the target object. TransRefer3D~\cite{he2021transrefer3d} uses an entity-aware attention module and a relation-aware attention module for fine-grained cross-modal matching. 3DJCG~\cite{cai20223djcg} devises a joint framework for 3D visual grounding~\cite{chen2020scanrefer} and 3D dense captioning~\cite{chen2021scan2cap} tasks, and their experiments demonstrate that extra caption-level data can improve the performance of 3D visual grounding.

In contrast to these supervised methods, our approach learns to localize target objects in 3D space using only caption-level annotations.

\begin{figure*}[htb]
\centering
\includegraphics[width=1\textwidth]{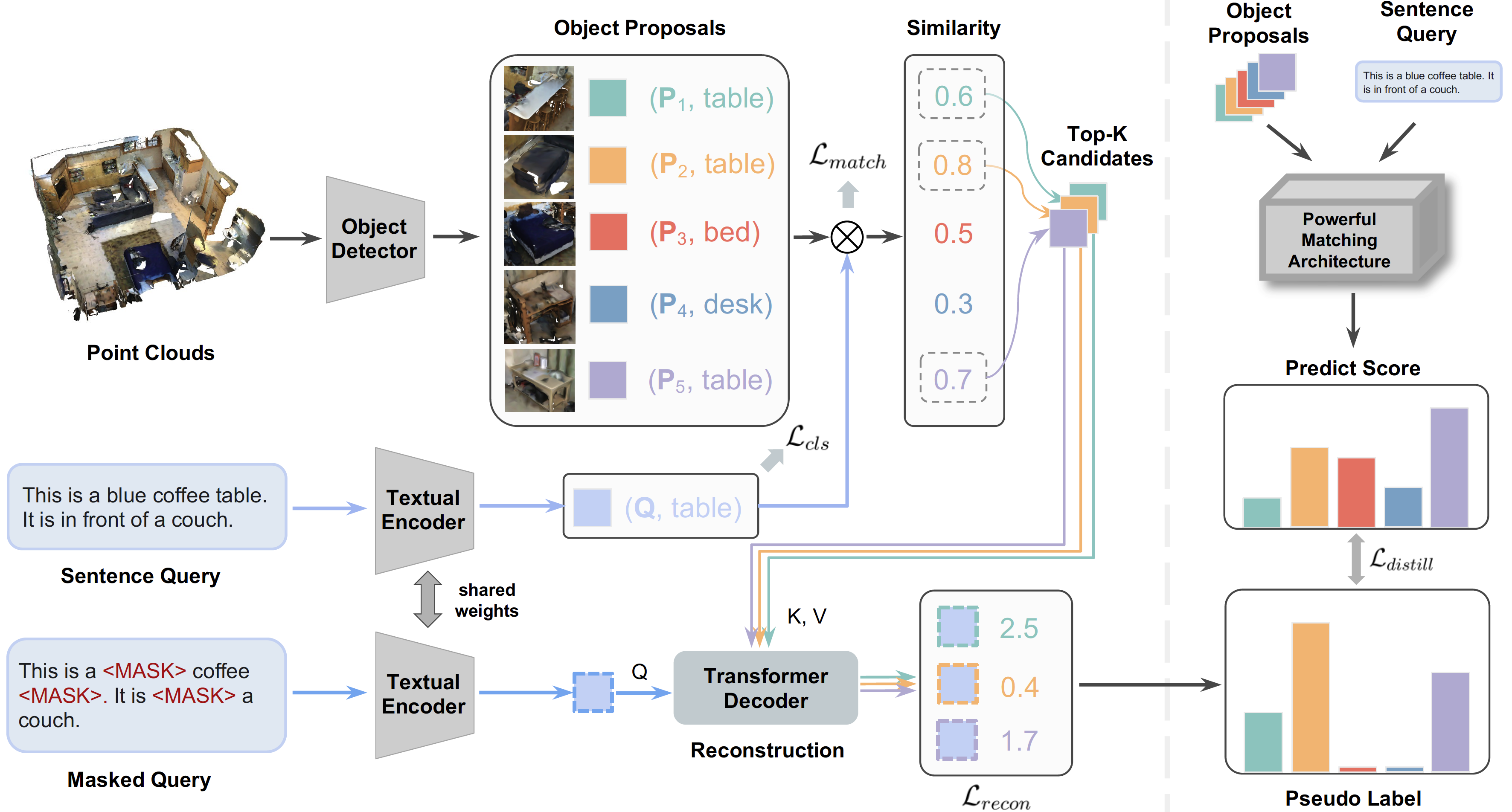}
\caption{Overall architecture diagram of our model. The model is based on a two-stage grounding pipeline. We first extract object proposals by pre-trained object detector. Then, we propose a coarse-to-fine semantic matching process to find the matched object-query pair. Furthermore, we distill the semantic matching knowledge into an effective matching architecture to enhance the inference efficiency.}
\label{fig:overall}
\end{figure*}

\noindent \textbf{Weakly Supervised Image Grounding. }
The image grounding task, similar to 3DVG, aims to identify objects in an image based on a sentence, and has a wide range of applications~\cite{plummer2015flickr30k, kazemzadeh2014referitgame, deng2021transvg, yu2018mattnet, kamath2021mdetr,yang2020improving}. Weakly supervised image grounding, which requires only images and corresponding sentences in the training phase, has gained popularity due to the low cost of annotation~\cite{gupta2020contrastive, rohrbach2016grounding, wang2021improving, faghri2017vse++, datta2019align2ground}.

Weakly supervised image grounding is typically treated as a Multiple Instance Learning (MIL) problem~\cite{ilse2018attention,maron1997framework}, where the image is represented as a bag of regions, generated by a pre-trained image object detector. Image-sentence matching scores are calculated based on region-phrase similarity scores, and ground-truth image-sentence links are used to supervise these scores. For example, ARN~\cite{liu2019adaptive} pairs image proposals and queries based on subject, location, and context information through adaptive grounding and collaborative reconstruction. InfoGround~\cite{gupta2020contrastive} proposes a contrastive learning objective function~\cite{he2020momentum} to optimize image-sentence scores. Wang et al.~\cite{wang2021improving} use a pre-trained image object detector to generate pseudo category labels for all regions, achieving region-phrase alignment by distilling knowledge from these pseudo labels.

However, MIL-based weakly supervised image grounding methods cannot solve the weakly supervised problem in 3DVG. Firstly, the presence of numerous different objects in a single 3D scene makes it difficult to learn a stable MIL classifier. Secondly, while image grounding aims to locate objects corresponding to all phrases in the sentence, 3DVG requires the identification of a single target object, necessitating a deeper and more comprehensive understanding of the sentence's semantic information, rather than just its phrases.

\section{Method}
\label{sec:method}

\subsection{Problem Formulation}
In this paper, we address the problem of weakly-supervised 3DVG. The input point cloud $\mathbf{P}=\{\mathbf{p}_i\}_{i=1}^{N_\mathrm{p}}$ contains point coordinates in 3D space, represented by $\mathbf{p}_i\in\mathbb{R}^{3}$. Correspondingly, a sentence query $\mathbf{Q}=\{\mathbf{q}_i\}_{i=1}^{N\mathrm{q}}$ is given to describe the object of interest. The objective of our model is to predict a 3D bounding box $\mathbf{B}=(\mathbf{c},\mathbf{r})$ that encompasses the object, where $\mathbf{c}=(c_x,c_y,c_z)$ represents the center of the box, and $\mathbf{r}=(r_x,r_y,r_z)$ represents the dimensions of the box. The number of input points and sentence length is denoted by $N_\mathrm{p}$ and $N_\mathrm{q}$, respectively. In the weakly-supervised setting, there are no bounding box annotations available during training.

\subsection{Overview}
As depicted in Figure \ref{fig:overall}, our model utilizes a two-stage grounding pipeline. In the first stage, we employ a pre-trained 3D object detector to extract $M_\mathrm{p}$ object proposals from the given point cloud. In the second stage, we propose a coarse-to-fine semantic matching process to evaluate the semantic similarity between each proposal and the sentence query. Specifically, the coarse-to-fine process comprises two steps. Firstly, we coarsely extract the top $K$ object proposals, which are referred to as candidates, by computing the object-sentence similarity matrix between all proposals and the sentence query. Secondly, we generate a more accurate pseudo label by considering the semantic reconstruction result of each candidate-sentence pair. Further details will be explained in Section~\ref{sec:method_coarse} and Section~\ref{sec:method_fine}.

Moreover, for reducing the inference costs and further enhancing the performance, we propose to distill the semantic matching knowledge into a supervised 3DVG pipeline as elaborated in Section~\ref{sec:method_distill}. Most advanced fully-supervised models typically operate using a ``detection-and-matching" paradigm. This means that these powerful matching architectures can be used as plug-and-play modules to incorporate knowledge learned from weak supervision.

\subsection{Coarse-grained Candidate Selection}
\label{sec:method_coarse}
\paragraph{Object-Sentence Similarity.}
Although we have extracted numerous high-quality object proposals from the pre-trained 3D object detector, identifying the best-matched proposal with the sentence query is still challenging. This is because a 3D scene may contain many different classes of objects, and the semantic spaces between objects and the sentence are not aligned. To overcome this challenge, we propose calculating a similarity matrix between the objects and the sentence based on both class and feature levels.

For the class level, we can obtain the object class from the pre-trained 3D object detector and the text class from a text classifier. For simplicity, we choose to train the text classifier from scratch and the classification loss $\mathcal{L}_{cls}$ is a simple cross-entropy loss. Considering that the object detector might be pre-trained on another dataset, the object class set and the text class set may be inconsistent. Therefore, before directly comparing the object proposals and the sentence, we need to transfer the object class prediction to the target text class. To achieve this, we propose using a class transform matrix  $\mathbf{M}^\mathrm{c}\in\mathbb{R}^{N_\mathrm{o}^\mathrm{c}\times N_\mathrm{q}^\mathrm{c}}$ for class alignment. The matrix is based on the cosine similarity between the GloVe embeddings of different class names. Here, $N_\mathrm{o}^\mathrm{c}$ and $N_\mathrm{q}^\mathrm{c}$ denote the number of object classes and the number of words in the sentence query, respectively.

For the feature level, we align the feature representations of the objects and the sentence query using a contrastive learning approach. Specifically, we pull the positive object-query pairs in the same scene closer and push the negative pairs further apart in the semantic space. To achieve this, all the object-query pairs in the same scene are considered as positive pairs $\mathbb{P}$, while those from different scenes are considered as negative pairs $\mathbb{N}$. The feature matching loss for object-sentence feature alignment can be computed by

\begin{equation}
\mathcal{L}_{match}=-\operatorname{log}\left(\frac{\sum\limits_{(\mathbf{p}, \mathbf{q})\in\mathbb{P}}e^{\phi(\mathbf{p},\mathbf{q})}}{\sum\limits_{(\mathbf{p}, \mathbf{q})\in\mathbb{P}}e^{\phi(\mathbf{p},\mathbf{q})}+
\sum\limits_{(\mathbf{p}^{'}, 
\mathbf{q})\in\mathbb{N}}e^{\phi(\mathbf{p}^{'},\mathbf{q})}}\right),
\end{equation}

\noindent
where $\mathbf{p}$ represents an object proposal and $\mathbf{q}$ a sentence query. $\phi$ is the feature similarity function, which is a dot product in our practice.

We get the object-sentence similarity $\mathbf{\hat{s}}\in\mathbb{R}^{M_\mathrm{p}}$ by
\begin{equation}
\mathbf{\hat{s}}=\phi(\mathbf{\tilde{P}}^{\mathrm{c}}\mathbf{M}^\mathrm{c},\mathbf{\tilde{Q}}^\mathrm{c})+\phi(\mathbf{\tilde{P}},\mathbf{\tilde{Q}}),
\end{equation}
where $\mathbf{\tilde{P}}\in\mathbb{R}^{M_\mathrm{p}\times d}$ / $\mathbf{\tilde{Q}}\in\mathbb{R}^{N_\mathrm{q}\times d}$ is the encoded object/sentence feature, and $\mathbf{\tilde{P}}^{\mathrm{c}}\in\mathbb{R}^{N_\mathrm{o}^\mathrm{c}}$ / $\mathbf{\tilde{Q}}^\mathrm{c}\in\mathbb{R}^{N_\mathrm{q}^\mathrm{c}}$ is the object/sentence class prediction. $\phi$ is a similarity function ($e.g.$, cosine similarity or dot product). $M_\mathrm{p}$ is the number of object proposals. $d$ is the hidden dimension.

\paragraph{Top-K Selection.}
According to the object-sentence similarity, we coarsely select the top $K$ candidates $\mathbf{\tilde{C}}\in\mathbb{R}^{K\times d}$ out of the $M_\mathrm{p}$ proposals $\mathbf{\tilde{P}}\in\mathbb{R}^{M_\mathrm{p}\times d}$, which can effectively filter out proposals that are significantly different from the semantics of the sentence.



\subsection{Fine-grained Semantic Matching}
\label{sec:method_fine}
Given the $K$ object candidates, we propose a semantic reconstruction module to measure fine-grained semantic similarity between the objects and the sentence query.

As depicted in Figure~\ref{fig:overall}, we mask important words in the sentence query, such as the target object (\textit{table}), its attribute (\textit{blue}), and its relation to other objects (\textit{in front of}) in the scene. We reconstruct the masked words with the assistance of each candidate, respectively. The candidate that provides the most useful semantic information to predict the keywords and contains the least amount of noise is expected to be the best match.

We encode the masked sentence query using a textual encoder, denoted as $\tilde{\mathbf{Q}}^\mathrm{m}\in\mathbb{R}^{N_\mathrm{q}\times d}$. For the $k$-th candidate $\tilde{\mathbf{c}}^k\in\mathbb{R}^{d}$, we obtain the cross-modal semantic representation $\mathbf{f}^k=\{\mathbf{f}_i^k\}_{i=1}^{N_\mathrm{q}}\in\mathbb{R}^{N_\mathrm{q}\times d}$ by a transformer decoder
\begin{equation}
    \mathbf{f}^k=\operatorname{Dec}(\tilde{\mathbf{Q}}^\mathrm{m},\tilde{\mathbf{c}}^k).
\end{equation}
To predict the masked words, we compute the energy distribution $\mathbf{e}^k=\{\mathbf{e}_i^k\}_{i=1}^{N_\mathrm{q}}\in\mathbb{R}^{N_\mathrm{q}\times d}$ over the vocabulary by
\begin{equation}
    \mathbf{e}_i^k=\mathbf{W}\mathbf{f}_i^k+\mathbf{b},
\end{equation}
where $\mathbf{e}_i^k\in\mathbb{R}^{N_\mathrm{v}}$ represents the energy distribution of the $i$-th predicted word, and $N_\mathrm{v}$ is the number of words in the vocabulary. $\mathbf{W}\in\mathbb{R}^{N_\mathrm{v}\times d}$ and $\mathbf{b}\in\mathbb{R}^{N_\mathrm{v}}$ are learnable parameters of a fully-connected layer.

Then, we use a reconstruction loss to train the semantic reconstruction module to effectively learn key information from the object context and predict the masked words. Specifically, the reconstruction can be computed as
\begin{equation}
\begin{aligned}
\mathcal{L}_{recon}^k=-\sum_{i \in N_\mathrm{mask}} \operatorname{log}p(\mathbf{q}_{i}|\mathbf{e}_i^k),
\end{aligned}
\end{equation}
where $N_\mathrm{mask}$ represents positions of masked words in the query and $\mathcal{L}_{recon}^k$ is the reconstruction loss for the $k$-th candidate $\tilde{\mathbf{c}}^k$. Then the total loss for all the $K$ candidates is $\mathcal{L}_{recon}=\sum_{k=1}^K\mathcal{L}_{recon}^k$.

\begin{table*}[htbp]
\centering
\caption{Performance comparison on ScanRefer. ``SUN'' and ``SCAN'' denotes that the 3D object detector is pretrained on SUN RGB-D\cite{song2015sun} or ScanNet\cite{dai2017scannet}, respectively. For the ``$R@n,IoU@m$" metric, $n\in\{1,3\}$ and $m\in\{0.25,0.5\}$.}
\label{table:scanrefer}
\begin{tabular}{c|c|cc|cc|cc|cc}
\toprule
\multirow{3}{*}{} &
  \multirow{3}{*}{Method} &
  \multicolumn{6}{c|}{R@3} &
  \multicolumn{2}{c}{R@1} \\ \cline{3-10} 
  &
   &
  \multicolumn{2}{c|}{} &
  \multicolumn{2}{c|}{} &
  \multicolumn{2}{c|}{} &
  \multicolumn{2}{c}{} \\ [-2ex]
 &
   &
  \multicolumn{2}{c|}{Unique} &
  \multicolumn{2}{c|}{Multiple} &
  \multicolumn{2}{c|}{Overall} &
  \multicolumn{2}{c}{Overall} \\
 &
   &
  $m$=0.25 &
  $m$=0.5 &
  $m$=0.25 &
  $m$=0.5 &
  $m$=0.25 &
  $m$=0.5 &
  $m$=0.25 &
  $m$=0.5 \\ \midrule
\multirow{5}{*}{\begin{tabular}[c]{@{}c@{}}SUN\end{tabular}} 
 &
  &
  &
  &
  &
  &
  &
  &
   &
  \\ [-2ex]
&
  Upper Bound &
  \textcolor{gray}{57.07} &
  \textcolor{gray}{35.28} &
  \textcolor{gray}{55.30} &
  \textcolor{gray}{35.29} &
  \textcolor{gray}{55.65} &
  \textcolor{gray}{35.29} &
  \textcolor{gray}{-} &
  \textcolor{gray}{-} \\
  &
  Random &
  15.88 &
  6.99 &
  7.38 &
  3.28 &
  9.03 &
  3.96 &
  3.66 &
  1.37 \\
 &
  MIL-Margin~\cite{faghri2017vse++} &
  19.94 &
  10.51 &
  10.18 &
  3.60 &
  12.07 &
  4.94 &
  6.80 &
  2.37 \\
 &
  MIL-NCE~\cite{gupta2020contrastive} &
  19.13 &
  10.95 &
  7.57 &
  3.56 &
  9.81 &
  5.00 &
  5.64 &
  2.69 \\
 &
  \textbf{Ours} &
\textbf{24.07} &
  \textbf{18.05}&
  \textbf{12.54}&
  \textbf{7.50}&
  \textbf{14.78}&
  \textbf{9.55}&
  \textbf{10.43}&
  \textbf{6.37}\\ 
  
\hline
\multirow{5}{*}{\begin{tabular}[c]{@{}c@{}}SCAN\end{tabular}} 
&
  &
  &
  &
  &
  &
  &
  &
   &
  \\ [-2ex]
  &
  Upper Bound &
  \textcolor{gray}{93.82} &
  \textcolor{gray}{77.02} &
  \textcolor{gray}{72.61} &
  \textcolor{gray}{58.01} &
  \textcolor{gray}{76.72} &
  \textcolor{gray}{61.70} &
  \textcolor{gray}{-} &
  \textcolor{gray}{-} \\   &
  Random &
  21.36 &
  14.25 &
  10.10 &
  7.15 &
  12.28 &
  8.53 &
  4.74 &
  3.32 \\
 &
  MIL-Margin~\cite{faghri2017vse++} &
  29.54 &
  22.49 &
  11.48 &
  8.04 &
  14.99 &
  10.84 &
  8.16 &
  5.66 \\
 &
  MIL-NCE~\cite{gupta2020contrastive} &
  48.94 &
  40.76 &
  17.41 &
  13.73 &
  23.53 &
  18.97 &
  18.95 &
  14.06 \\
  &
  \textbf{Ours} &
\textbf{70.84} &
  \textbf{58.21} &
  \textbf{25.28}&
  \textbf{20.68}&
  \textbf{34.12}&
  \textbf{27.97}&
  \textbf{27.37}&
 \textbf{21.96}\\

  \bottomrule
\end{tabular}
\end{table*}

\subsection{Knowledge Distillation}
\label{sec:method_distill}
As mentioned earlier, a lower reconstruction loss indicates that the object candidate provides more consistent semantic information. A direct approach for object prediction is to select the candidate with the lowest reconstruction loss, as it is likely to be the best match. However, this coarse-to-fine matching process is computationally expensive during inference and not explicitly optimized for grounding tasks. To tackle the issues, we propose to distill the coarse-to-fine semantic matching knowledge into a supervised 3DVG pipeline. Our approach offers multiple benefits, including reduced inference costs and the ability to capitalize on more powerful 3DVG architectures and established learning objectives tailored for 3DVG tasks. By incorporating knowledge distillation, our framework can be integrated with any advanced supervised 3DVG pipeline, enhancing the flexibility and practicality of our method.

For candidates, we calculate the reward according to their rank of $\mathcal{L}_{recon}^k$. The reward is reduced from one to zero, under the assumption that lower reconstruction loss gets better reward. The distilled pseudo labels $\mathbf{d} = \{ d_1, ..., d_{M_\mathrm{p}} \}$ can be generated by filling the rewards of candidates to their original indices and padding the non-candidate indices with zeros, following by a SoftMax operation. After all, we distill the knowledge by aligning the predict scores $\mathbf{s} = \{ s_1, ..., s_{M_\mathrm{p}} \}$ to the pseudo labels, where the predict scores are obtained from the powerful matching architecture. The distillation loss is:
\begin{equation}
\mathcal{L}_{distill}=-\sum_{i=1}^{M_\mathrm{p}}d_i\operatorname{log}(\frac{\operatorname{exp}(s_i)}{\sum_{j=1}^{M_\mathrm{p}}\operatorname{exp}(s_j)}).
\end{equation}

\subsection{Training and Inference}
\label{sec:model_train}

\paragraph{Multi-Task Loss}
We train the model end-to-end via a multi-task loss function, formulated by
\begin{equation}
\mathcal{L}_{overall}=\mathcal{L}_{distill}+\lambda_1\mathcal{L}_{cls}+\lambda_2\mathcal{L}_{match}+\lambda_3\mathcal{L}_{recon}
\end{equation}
where $\lambda_1$, $\lambda_2$ and $\lambda_3$ are hyper-parameters to balance four parts of the loss function.

\paragraph{Inference.}
Thanks to the knowledge distillation, all we need in the inference phase is the two-stage 3DVG pipeline. We get the predict score $\mathbf{s}\in\mathbb{R}^{M_\mathrm{p}}$ from the matching architecture, and the index of the predicted best-match proposal is $\operatorname{argmax}(\mathbf{s})$. Then, we obtain the corresponding 3D bounding box of this object proposal.

\begin{table*}[htbp]
\centering
\tabcolsep=1.5mm
\caption{Performance comparison on Nr3D and Sr3D dataset. ``SUN'' and ``SCAN'' denotes that the 3D object detector is pretrained on SUN RGB-D~\cite{song2015sun} or ScanNet~\cite{dai2017scannet}, respectively. For the ``$R@n,IoU@m$" metric, $n=3$ and $m\in\{0.25,0.5\}$.}
\label{table:referit}
\resizebox{0.9\textwidth}{!}{
\begin{tabular}{cccccccccccc}
\toprule
\multicolumn{1}{c|}{\multirow{2}{*}{}} &
  \multicolumn{1}{c|}{\multirow{2}{*}{Method}} &
  \multicolumn{2}{c|}{Easy} &
  \multicolumn{2}{c|}{Hard} &
  \multicolumn{2}{c|}{View-dep.} &
  \multicolumn{2}{c|}{View-indep.} &
  \multicolumn{2}{c}{Overall} \\
\multicolumn{1}{c|}{} &
  \multicolumn{1}{c|}{} &
  $m$=0.25 &
  \multicolumn{1}{c|}{$m$=0.5} &
  $m$=0.25 &
  \multicolumn{1}{c|}{$m$=0.5} &
  $m$=0.25 &
  \multicolumn{1}{c|}{$m$=0.5} &
  $m$=0.25 &
  \multicolumn{1}{c|}{$m$=0.5} &
  $m$=0.25 &
  $m$=0.5 \\ \midrule
\multicolumn{12}{c}{\textbf{Nr3D}} \\ \hline
\multicolumn{1}{c|}{\multirow{5}{*}{SUN}} &
\multicolumn{1}{c|}{} &
 &
  \multicolumn{1}{c|}{} &
   &
  \multicolumn{1}{c|}{} &
   &
  \multicolumn{1}{c|}{} &
  &
  \multicolumn{1}{c|}{} &
   &
   \\ [-2ex]
\multicolumn{1}{c|}{} &
  \multicolumn{1}{c|}{Upper Bound} &
  \textcolor{gray}{40.24} &
  \multicolumn{1}{c|}{\textcolor{gray}{24.62}} &
  \textcolor{gray}{40.62} &
  \multicolumn{1}{c|}{\textcolor{gray}{23.80}} &
  \textcolor{gray}{40.66} &
  \multicolumn{1}{c|}{\textcolor{gray}{24.88}} &
  \textcolor{gray}{40.32} &
  \multicolumn{1}{c|}{\textcolor{gray}{23.82}} &
  \textcolor{gray}{40.44} &
  \textcolor{gray}{24.20} \\
\multicolumn{1}{c|}{} &
  \multicolumn{1}{c|}{Random} &
  6.70 &
  \multicolumn{1}{c|}{2.40} &
  6.34 &
  \multicolumn{1}{c|}{2.75} &
  6.59 &
  \multicolumn{1}{c|}{2.91} &
  6.47 &
  \multicolumn{1}{c|}{2.41} &
  6.51 &
  2.59 \\
\multicolumn{1}{c|}{} &
  \multicolumn{1}{c|}{MIL-Margin~\cite{faghri2017vse++}} &
  9.93 &
  \multicolumn{1}{c|}{5.63} &
  7.79 &
  \multicolumn{1}{c|}{4.03} &
  8.71 &
  \multicolumn{1}{c|}{4.77} &
  8.88 &
  \multicolumn{1}{c|}{4.81} &
  8,82 &
  4,80 \\
\multicolumn{1}{c|}{} &
  \multicolumn{1}{c|}{MIL-NCE~\cite{gupta2020contrastive}} &
  9.93 &
  \multicolumn{1}{c|}{5.42} &
  7.77 &
  \multicolumn{1}{c|}{4.79} &
  8.45 &
  \multicolumn{1}{c|}{4.67} &
  9.00 &
  \multicolumn{1}{c|}{5.32} &
  8.81 &
  5.09 \\
\multicolumn{1}{c|}{} &
  \multicolumn{1}{c|}{\textbf{Ours}} &
  \textbf{10.93} &
  \multicolumn{1}{c|}{\textbf{6.36}} &
  \textbf{9.83} &
  \multicolumn{1}{c|}{\textbf{6.18}} &
  \textbf{10.77} &
  \multicolumn{1}{c|}{\textbf{6.53}} &
  \textbf{10.13} &
  \multicolumn{1}{c|}{\textbf{6.13}} &
  \textbf{10.36} &
  \textbf{6.27} \\ \hline
\multicolumn{1}{c|}{\multirow{5}{*}{SCAN}} &
\multicolumn{1}{c|}{} &
 &
  \multicolumn{1}{c|}{} &
   &
  \multicolumn{1}{c|}{} &
   &
  \multicolumn{1}{c|}{} &
  &
  \multicolumn{1}{c|}{} &
   &
   \\ [-2ex]
\multicolumn{1}{c|}{} &
  \multicolumn{1}{c|}{Upper Bound} &
  \textcolor{gray}{62.43} &
  \multicolumn{1}{c|}{\textcolor{gray}{44.75}} &
  \textcolor{gray}{58.98} &
  \multicolumn{1}{c|}{\textcolor{gray}{44.18}} &
  \textcolor{gray}{59.15} &
  \multicolumn{1}{c|}{\textcolor{gray}{42.91}} &
  \textcolor{gray}{61.44} &
  \multicolumn{1}{c|}{\textcolor{gray}{45.29}} &
  \textcolor{gray}{60.64} &
  \textcolor{gray}{44.45} \\
\multicolumn{1}{c|}{} &
  \multicolumn{1}{c|}{Random} &
  8.81 &
  \multicolumn{1}{c|}{5.66} &
  7.57 &
  \multicolumn{1}{c|}{4.97} &
  7.28 &
  \multicolumn{1}{c|}{4.80} &
  8.65 &
  \multicolumn{1}{c|}{5.61} &
  8.17 &
  5.30 \\
\multicolumn{1}{c|}{} &
  \multicolumn{1}{c|}{MIL-Margin~\cite{faghri2017vse++}} &
  14.25 &
  \multicolumn{1}{c|}{10.64} &
  9.79 &
  \multicolumn{1}{c|}{7.68} &
  10.64 &
  \multicolumn{1}{c|}{8.35} &
  12.63 &
  \multicolumn{1}{c|}{9.50} &
  11.93 &
  9.10 \\
\multicolumn{1}{c|}{} &
  \multicolumn{1}{c|}{MIL-NCE~\cite{gupta2020contrastive}} &
  17.29 &
  \multicolumn{1}{c|}{13.53} &
  9.61 &
  \multicolumn{1}{c|}{7.59} &
  11.96 &
  \multicolumn{1}{c|}{9.44} &
  14.01 &
  \multicolumn{1}{c|}{10.98} &
  13.29 &
  10.44 \\
\multicolumn{1}{c|}{} &
  \multicolumn{1}{c|}{\textbf{Ours}} &
  \textbf{27.29} &
  \multicolumn{1}{c|}{\textbf{21.10}} &
  \textbf{17.98} &
  \multicolumn{1}{c|}{\textbf{14.42}} &
  \textbf{21.60} &
  \multicolumn{1}{c|}{\textbf{16.80}} &
  \textbf{22.91} &
  \multicolumn{1}{c|}{\textbf{18.07}} &
  \textbf{22.45} &
  \textbf{17.62} \\ \midrule
\multicolumn{12}{c}{\textbf{Sr3D}} \\ \hline
\multicolumn{1}{c|}{\multirow{5}{*}{SUN}} &
\multicolumn{1}{c|}{} &
 &
  \multicolumn{1}{c|}{} &
   &
  \multicolumn{1}{c|}{} &
   &
  \multicolumn{1}{c|}{} &
  &
  \multicolumn{1}{c|}{} &
   &
   \\ [-2ex]
\multicolumn{1}{c|}{} &
  \multicolumn{1}{c|}{Upper Bound} &
  \textcolor{gray}{39.22} &
  \multicolumn{1}{c|}{\textcolor{gray}{23.69}} &
  \textcolor{gray}{39.58} &
  \multicolumn{1}{c|}{\textcolor{gray}{21.83}} &
  \textcolor{gray}{25.93} &
  \multicolumn{1}{c|}{\textcolor{gray}{13.30}} &
  \textcolor{gray}{39.92} &
  \multicolumn{1}{c|}{\textcolor{gray}{23.24}} &
  \textcolor{gray}{39.33} &
  \textcolor{gray}{22.82} \\
\multicolumn{1}{c|}{} &
  \multicolumn{1}{c|}{Random} &
  6.53 &
  \multicolumn{1}{c|}{2.28} &
  4.61 &
  \multicolumn{1}{c|}{2.17} &
  1.86 &
  \multicolumn{1}{c|}{0.80} &
  6.05 &
  \multicolumn{1}{c|}{2.32} &
  5.96 &
  2.25 \\
\multicolumn{1}{c|}{} &
  \multicolumn{1}{c|}{MIL-Margin~\cite{faghri2017vse++}} &
  8.52 &
  \multicolumn{1}{c|}{4.84} &
  5.66 &
  \multicolumn{1}{c|}{3.98} &
  3.19 &
  \multicolumn{1}{c|}{\textbf{2.66}} &
  7.86 &
  \multicolumn{1}{c|}{4.67} &
  7.67 &
  4.59 \\
\multicolumn{1}{c|}{} &
  \multicolumn{1}{c|}{MIL-NCE~\cite{gupta2020contrastive}} &
  8.66 &
  \multicolumn{1}{c|}{4.92} &
  4.10 &
  \multicolumn{1}{c|}{2.78} &
  2.46 &
  \multicolumn{1}{c|}{0.93} &
  7.56 &
  \multicolumn{1}{c|}{4.42} &
  7.30 &
  4.28 \\
\multicolumn{1}{c|}{} &
  \multicolumn{1}{c|}{\textbf{Ours}} &
  \textbf{10.31} &
  \multicolumn{1}{c|}{\textbf{6.60}} &
  \textbf{8.57} &
  \multicolumn{1}{c|}{\textbf{6.23}} &
  \textbf{4.19} &
  \multicolumn{1}{c|}{1.86} &
  \textbf{10.09} &
  \multicolumn{1}{c|}{\textbf{6.69}} &
  \textbf{9.79} &
  \textbf{6.49} \\ \hline
\multicolumn{1}{c|}{\multirow{5}{*}{SCAN}} &
\multicolumn{1}{c|}{} &
 &
  \multicolumn{1}{c|}{} &
   &
  \multicolumn{1}{c|}{} &
   &
  \multicolumn{1}{c|}{} &
  &
  \multicolumn{1}{c|}{} &
   &
   \\ [-2ex]
\multicolumn{1}{c|}{} &
  \multicolumn{1}{c|}{Upper Bound} &
  \textcolor{gray}{65.42} &
  \multicolumn{1}{c|}{\textcolor{gray}{46.75}} &
  \textcolor{gray}{58.46} &
  \multicolumn{1}{c|}{\textcolor{gray}{42.69}} &
  \textcolor{gray}{53.59} &
  \multicolumn{1}{c|}{\textcolor{gray}{34.84}} &
  \textcolor{gray}{63.77} &
  \multicolumn{1}{c|}{\textcolor{gray}{46.01}} &
  \textcolor{gray}{63.34} &
  \textcolor{gray}{45.54} \\
\multicolumn{1}{c|}{} &
  \multicolumn{1}{c|}{Random} &
  8.50 &
  \multicolumn{1}{c|}{5.38} &
  6.85 &
  \multicolumn{1}{c|}{4.55} &
  5.59 &
  \multicolumn{1}{c|}{3.72} &
  8.12 &
  \multicolumn{1}{c|}{5.20} &
  8.01 &
  5.13 \\
\multicolumn{1}{c|}{} &
  \multicolumn{1}{c|}{MIL-Margin~\cite{faghri2017vse++}} &
  12.55 &
  \multicolumn{1}{c|}{9.82} &
  9.59 &
  \multicolumn{1}{c|}{7.50} &
  9.57 &
  \multicolumn{1}{c|}{7.98} &
  11.76 &
  \multicolumn{1}{c|}{9.18} &
  11.67 &
  9.13 \\
\multicolumn{1}{c|}{} &
  \multicolumn{1}{c|}{MIL-NCE~\cite{gupta2020contrastive}} &
  17.45 &
  \multicolumn{1}{c|}{12.51} &
  9.61 &
  \multicolumn{1}{c|}{7.14} &
  12.37 &
  \multicolumn{1}{c|}{7.97} &
  15.22 &
  \multicolumn{1}{c|}{11.03} &
  15.11 &
  10.90 \\
\multicolumn{1}{c|}{} &
  \multicolumn{1}{c|}{\textbf{Ours}} &
  \textbf{29.40} &
  \multicolumn{1}{c|}{\textbf{24.87}} &
  \textbf{21.00} &
  \multicolumn{1}{c|}{\textbf{17.47}} &
  \textbf{20.21} &
  \multicolumn{1}{c|}{\textbf{17.15}} &
  \textbf{27.19} &
  \multicolumn{1}{c|}{\textbf{22.90}} &
  \textbf{26.89} &
  \textbf{22.66} \\ \bottomrule
\end{tabular}}
\end{table*}

\section{Experiments}
\label{sec:exp}

\subsection{Datasets}
\noindent \textbf{ScanRefer.} The ScanRefer~\cite{chen2020scanrefer} dataset contain 51,583 descriptions of 11,046 objects from 800 ScanNet~\cite{dai2017scannet} scenes. On average, each scene has 64.48 sentences and 13.81 objects. The data can be divided into ``Unique" and ``Multiple", depending on whether there are multiple objects of the same category as the target in the scene.

\noindent \textbf{Nr3D/Sr3D.} The Nr3D/Sr3D dataset~\cite{achlioptas2020referit3d} is also based on the 3D scene dataset ScanNet~\cite{dai2017scannet}. Nr3D contains 41,503 human utterances collected by ReferItGame, and Sr3D contains 83,572 sentences automatically generated based on a ``target-spatial relationship-anchor object" template. Similar to the definition of ``Unique" and ``Multiple" in ScanRefer, Nr3D/Sr3D can be split into ``easy" and ``hard" subsets. The ``view-dep." and ``view-indep." subsets depend on whether the description is related to the speaker's view. \footnote{In the Nr3D/Sr3D datasets, the supervised task involves selecting the correct matching 3D box from a set of given boxes, with the instance matching accuracy serving as the evaluation metric. However, in the weakly-supervised setting, we predict the boxes from scratch and assess the IoU metrics, which cannot be directly compared to the results of supervised methods.}

\subsection{Evaluation Metric.} To evaluate the performance of our method and baselines on these three datasets, we adopt the ``$R@n, IoU@m$" metric. Specifically, this metric represents the percentage of at least one of the top-$n$ predicted proposals having an IoU greater than $m$ when compared to the ground-truth target bounding box. In our setting, $n \in {1, 3}$ and $m \in {0.25, 0.5}$. 

\subsection{Implementation Details.} In our practice, we use the pretrained GroupFree model~\cite{liu2021group} as our 3D object detector and distill the learned semantic matching knowledge to the matching architecture proposed in 3DJCG~\cite{cai20223djcg}. The input point number $N_\mathrm{p}$, the proposal number $M_\mathrm{p}$, and the candidate number $K$ are set to 50000, 256 and 8, respectively. More details can be found in the supplementary material.

\subsection{Compared Methods}

\noindent \textbf{Random.} We randomly select a candidate from all the proposals as the predicted result.

\noindent \textbf{MIL-Margin.} The MIL-Margin method~\cite{faghri2017vse++} proposes a max margin loss to enforce the score between a sentence and a paired scene to be higher than non-paired scenes, and vice versa. 

\noindent \textbf{MIL-NCE.} The MIL-NCE method~\cite{gupta2020contrastive} maximizes the InfoNCE lower bound on mutual information between the sentence and proposals from the paired scene, compared to non-corresponding pairs of scenes and sentences.

\noindent \textbf{Upper Bound.} The quality of the bounding boxes generated by the 3D object detector determines the upper bound performance of our model. We consider the maximum IoU between all the $M_\mathrm{p}$ object proposals and the ground-truth bounding box as the upper bound.

\subsection{Quantitative Comparison}
The performance results of our methods and baselines on ScanRefer and Nr3D/Sr3D are reported in Table \ref{table:scanrefer} and Table \ref{table:referit}, respectively, with the best results highlighted in \textbf{bold}. The comparison to supervised methods is presented in Table \ref{table:sup}. Although the 3D object detector pre-trained on ScanNet implicitly utilizes ground truth boxes on ScanNet, the object-sentence annotations are still unseen, and pre-training on ScanNet is only used to obtain more accurate proposals. To fully avoid annotations in ScanNet, we also evaluate results using a detector pre-trained on SUN RGB-D~\cite{song2015sun}. Despite the degradation caused by out-of-domain data, our method still shows significant improvement over baselines. By analyzing the evaluation results, we can observe the following facts:
\begin{itemize}

\item Our method achieves significant improvements over the Random method on all datasets, indicating the effectiveness of the coarse-to-fine semantic matching model in analyzing the similarity between objects and sentences when true object-sentence pairs are unavailable.
\item The results show that our method outperforms widely used MIL-based weakly supervised methods by a large margin, and even approaches the upper bound in the ``Unique" subset of ScanRefer. This suggests that our proposed model can deeply exploit the alignment relationship between 3D scenes and sentences and identify the most semantically relevant object proposals.
\item Our coarse-to-fine semantic matching model significantly improves performance in the challenging ``Multiple" subset of ScanRefer and ``Hard" subset of Nr3D/Sr3D, where there are multiple interfering objects with the same category as the target object. This problem requires a comprehensive understanding of the sentence to distinguish the described object, which our model handles efficiently with the keywords semantic reconstruction module.
\item The performance improvement with the SUN RGB-D pre-trained backbone is relatively small on Nr3D and Sr3D datasets, possibly because the target objects are inherently more challenging to detect, and the pre-trained detector performs poorly due to out-of-distribution data. The low grounding upper bound and inaccurate proposals make the training phase unstable. Nevertheless, our method outperforms all baselines, and when the detector is more reliable, our semantic matching model shows much more significant advantages on Nr3D/Sr3D.

\end{itemize}

\begin{table}[tbp]
    \centering
    \caption{Comparison to supervised methods on ScanRefer.}
    \label{table:sup}
    \scalebox{0.85}{
    \begin{tabular}{c|c|cc}
    \toprule
    \multirow{2}{*}{Method} & \multirow{2}{*}{Backbone} & \multicolumn{2}{c}{R@1} \\ \cline{3-4} 
                        &    &             &           \\ [-2ex]
                        &    & m=0.25    & m=0.5   \\ \midrule
    ScanRefer~\cite{chen2020scanrefer}  & VoteNet             & 41.19       & 27.40     \\
    SAT~\cite{yang2021sat}    & VoteNet       & 44.54       & 30.14     \\
    3DVG-Transformer~\cite{zhao20213dvg} & VoteNet       & 47.57       & 34.67     \\
    3DJCG~\cite{cai20223djcg}   & VoteNet                & 49.56       & 37.33     \\ \hline
    & & & \\ [-2ex]
    Ours         & VoteNet             & 25.87       & 16.63     \\
    Ours         & GroupFree           & \textbf{27.37}       & \textbf{21.96}     \\ \bottomrule
    \end{tabular}}
    \end{table}
\begin{table}[t]
\centering
\tabcolsep=1.8mm
\caption{Ablation studies on the coarse-to-fine semantic matching model. The experiments of all the ablation study are conducted on ScanRefer dataset. ``R@3, SUN'' refers to $n=3$ and the object detector is pretrained on SUN RGB-D.}
\label{table:abalation}
\scalebox{0.85}{
\begin{tabular}{ccc|cc|cc}
\toprule
\multirow{2}{*}{$\mathcal{L}_{cls}$} & \multirow{2}{*}{$\mathcal{L}_{match}$} & \multirow{2}{*}{$\mathcal{L}_{recon}$} & \multicolumn{2}{c|}{R@3, SUN} & \multicolumn{2}{c}{R@3, SCAN} \\ \cline{4-7}
&               &               & &  & &  \\ [-2ex]
          &               &               & $m$=0.25 & $m$=0.5 & $m$=0.25 & $m$=0.5 \\ \midrule
            &               &               & 9.03     & 3.96    & 12.28    & 8.53    \\
            & \checkmark    &               & 10.53    & 6.29    & 17.41    & 14.28   \\
\checkmark  &               &               & 13.10    & 8.17    & 32.71    & 25.90   \\
\checkmark  & \checkmark    &               & 13.76    & 8.48    & 33.03    & 26.58   \\
\checkmark  & \checkmark    & \checkmark    & \textbf{14.78}    & \textbf{9.55}    & \textbf{34.12}    & \textbf{27.97}   \\ \bottomrule
\end{tabular}
}
\end{table}

\subsection{Ablation Study}
To further assess the effectiveness of each component, we conduct ablation studies on the ScanRefer dataset.

\subsubsection{Effectiveness of Semantic Matching Model}
\noindent \textbf{Coarse-to-Fine Matching Scores.}
We aim to examine the effect of each module in the coarse-to-fine semantic matching model. $\mathcal{L}_{cls}$ and $\mathcal{L}_{match}$ denote whether to use class similarity and feature similarity for coarsely selecting the top-$K$ candidates, respectively. $\mathcal{L}_{recon}$ represents using the reconstruct module to finely generate distilled pseudo labels for the selected $K$ candidates. If $\mathcal{L}_{recon}$ is not used, all the selected $K$ candidates's rewards are set directly to 1. Table \ref{table:abalation} shows that using $\mathcal{L}_{cls}$ or $\mathcal{L}_{match}$ alone can effectively aid the model in learning object-sentence pairs from the caption-level annotations, while joint usage of $\mathcal{L}_{cls}$ and $\mathcal{L}_{match}$ leads to better performance. Furthermore, the last two rows suggest that the fine-grained semantic matching module can effectively and comprehensively analyze the semantic similarity between the selected $K$ candidates and the sentence query, and further enhance the performance.

\noindent \textbf{Number of Coarse-grain Candidate.} 
We analyze the performance for varying numbers of coarse-grained candidates, $K \in {4, 8, 16, 32}$. As shown in Table \ref{table:num_can}, we observe that selecting 8 candidates yields the best results for fine-grained semantic matching. We tentatively infer the reason is that too small $K$ leaves out the possible proposal that covers the target object, while too large $K$ leads to selecing many proposals that are not relevant to the description due to the numerous objects in a 3D scene.

\begin{table}[t]
\centering
\caption{Ablation study on the candidate number $K$.}
\label{table:num_can}
\scalebox{0.85}{
\begin{tabular}{c|cc|cc}
\toprule
\multirow{2}{*}{$K$} & \multicolumn{2}{c|}{R@3, SUN} & \multicolumn{2}{c}{R@3, SCAN} \\ \cline{2-5} 
& & & & \\ [-2ex]
   & $m$=0.25 & $m$=0.5 & $m$=0.25 & $m$=0.5 \\ \midrule
4  & 14.04    & 9.05    & 33.68    & 27.50   \\
\textbf{8}                                & \textbf{14.78}    & \textbf{9.55}   & \textbf{34.12}  & \textbf{27.97} \\
16 & 14.37    & 9.39    & 33.23    & 26.99   \\
32 & 14.26    & 9.21    & 31.29    & 25.90   \\ \bottomrule
\end{tabular}}
\end{table}
\begin{table}[t]
\centering
\tabcolsep=0.8mm
\caption{Ablation study on the knowledge distillation. The matching time is evaluated for one batch.}
\label{table:distill}
\scalebox{0.85}{
\begin{tabular}{c|cc|cc|cc}
\toprule
          \multirow{2}{*}{Distill Target} & \multicolumn{2}{c|}{R@3, SUN} & \multicolumn{2}{c|}{R@3, SCAN} & \multicolumn{2}{c}{Matching Phase} \\ \cline{2-7} 
           & & & & & & \\[-2ex]
            & $m$=0.25 & $m$=0.5 & $m$=0.25 & $m$=0.5 & Time & Params \\ \midrule
w/o distill. & 13.88    & 9.09    & 31.71    & 26.38   & 31.4 ms    & 5.85 M    \\
SAT~\cite{yang2021sat} & 14.00    & 9.15    & 33.70    & 27.85   & 6.78 ms    & 1.85 M    \\
3DVG-Trans~\cite{zhao20213dvg} & 14.01    & 9.10    & \textbf{34.61}    & \textbf{28.22}  & 8.38 ms     & 1.91 M    \\
3DJCG~\cite{cai20223djcg} & \textbf{14.78}    & \textbf{9.55}   & 34.12  & 27.97  & 8.22 ms    & 1.93 M   \\ \bottomrule
\end{tabular}
}
\end{table}

\subsubsection{Effectiveness of Knowledge Distillation}
\label{exp:distill}

\begin{figure*}[tb]
    \centering
    \includegraphics[width=1\textwidth]{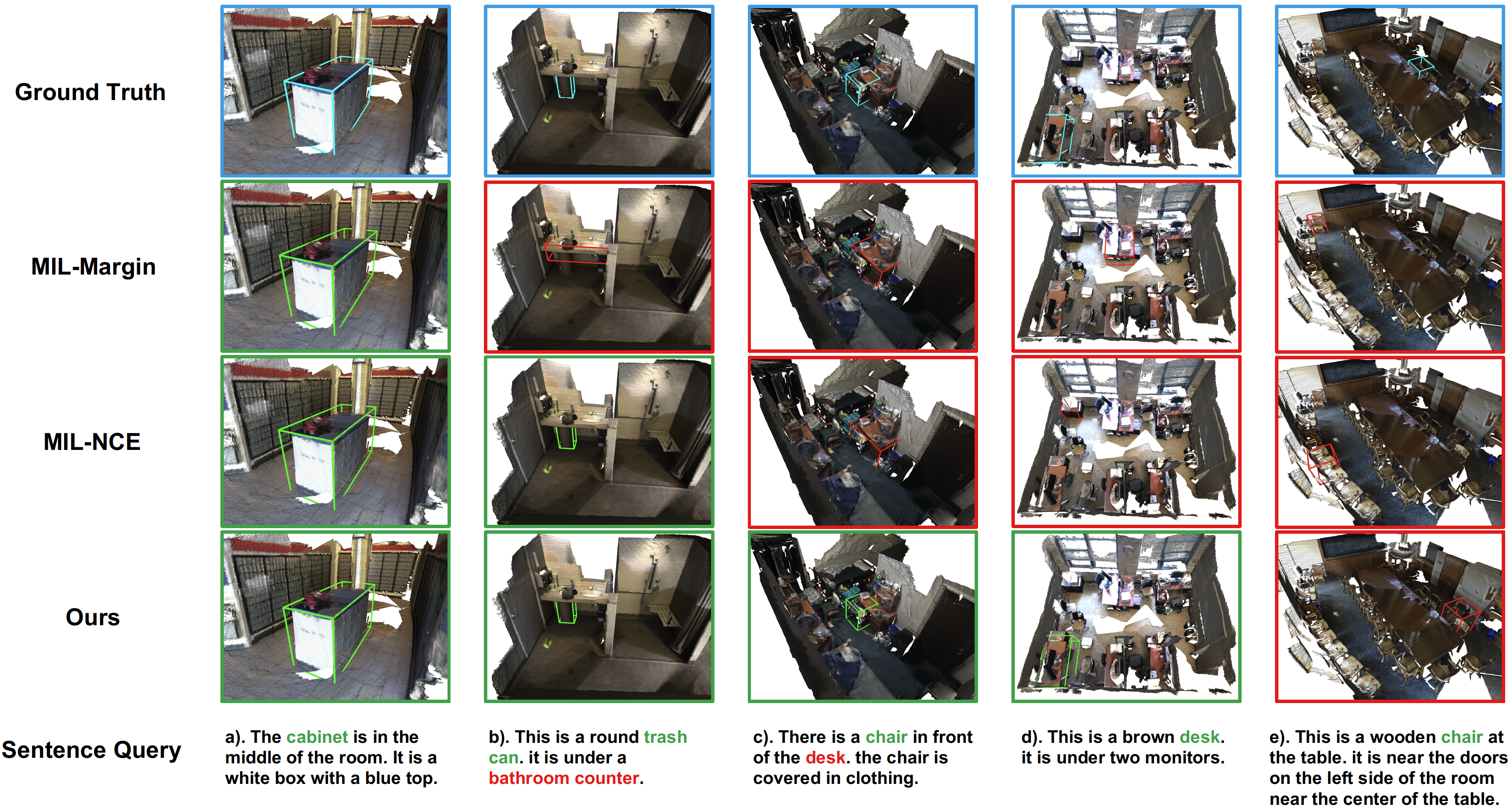}
    \caption{Qualitative Comparison between MIL-based methods and Ours. }
    \label{fig:quali_1}
\end{figure*}

To investigate the effect of the semantic distillation in terms of performance and efficiency, we construct the baseline without knowledge distillation that removes the distillation loss $\mathcal{L}_{distill}$ during training and directly uses the coarse-to-fine semantic matching model for inference. As shown in Table \ref{table:distill}, distilling the semantic matching knowledge into the matching module of a two-stage 3DVG model brings a significant performance improvement. The well-studied structure of the existing 3DVG model enhances the generalization ability of our method. As for the efficiency, we observe that the distilled matching module is 3$\times$ smaller and 4$\times$ faster than the coarse-to-fine semantic matching model, demonstrating that the distilling operation reduces inference costs significantly. Meanwhile, we try to distill the knowledge into the matching modules of different supervised methods (SAT~\cite{yang2021sat}, 3DVG-Transformer~\cite{zhao20213dvg}, and 3DJCG~\cite{cai20223djcg}), the results show that the distillation fits well to different architectures.

\subsection{Qualitative Comparison}
As depicted in Figure~\ref{fig:quali_1}, we visualize the predicted bounding boxes in the corresponding 3D scene, where the green box denotes that the method predicts the correct object (IoU $\geq 0.5$ with the true box), and the red box indicates a wrong prediction.

In case (a), the target object \textit{cabinet} is the only cabinet in the simple scene. So, both MIL-based methods and our method can predict the box well. In case (b) / (c), the target object is a \textit{trash can} / \textit{chair}. MIL-based methods may be misled by the presence of another object (\textit{bathroom counter} / \textit{desk}) in the sentence. While our method can filter out the objects that do not belong to the target category, benefiting from the coarse-grained candidate selection module. In case (d), there are six \textit{desks} in the scene. The MIL-based methods fail to localize the correct object, even though they figure out the target object (category) is \textit{desk}. With the fine-grained semantic matching module, our methods can better differentiate among these six \textit{desks} and choose the one best-matched to the sentence (``brown'' and ``under two monitors''). In case (e), the scene contains 32 different \textit{chairs}. Unfortunately, both our method and MIL-based methods fail in this case. However, we consider our method's predicted result acceptable. Firstly, the sentence query's expressions, such as ``near the doors'' and ``near the center'', are ambiguous and cannot give a precise location of the target object. Secondly, our method's predicted \textit{chair} is also consistent with the sentence description and is close to the true \textit{chair}.

\section{Conclusion}
In this paper, we raise the weakly-supervised 3D visual grounding setting, using only coarse scene-sentence correspondences to learn the object-sentence links. The weak supervision gets rid of time-consuming and expensive manual annotations of accurate bounding boxes, which makes this problem more realistic but more challenging. To tackle this, we propose a novel semantic matching method to analyze the object-sentence semantic similarity in a coarse-to-fine manner. Moreover, we distill the semantic matching knowledge into the existing 3D visual grounding architecture, effectively reducing the inference cost and further improving performance. The sufficient experiments on large-scale datasets verify the effectiveness of our method.

{\small
\bibliographystyle{ieee_fullname}
\bibliography{main}

\begin{thebibliography}{10}\itemsep=-1pt

\bibitem{achlioptas2020referit3d}
Panos Achlioptas, Ahmed Abdelreheem, Fei Xia, Mohamed Elhoseiny, and Leonidas
  Guibas.
\newblock Referit3d: Neural listeners for fine-grained 3d object identification
  in real-world scenes.
\newblock In {\em European Conference on Computer Vision}, pages 422--440.
  Springer, 2020.

\bibitem{bird2009natural}
Steven Bird, Ewan Klein, and Edward Loper.
\newblock {\em Natural language processing with Python: analyzing text with the
  natural language toolkit}.
\newblock " O'Reilly Media, Inc.", 2009.

\bibitem{cai20223djcg}
Daigang Cai, Lichen Zhao, Jing Zhang, Lu Sheng, and Dong Xu.
\newblock 3djcg: A unified framework for joint dense captioning and visual
  grounding on 3d point clouds.
\newblock In {\em Proceedings of the IEEE/CVF Conference on Computer Vision and
  Pattern Recognition}, pages 16464--16473, 2022.

\bibitem{chen2020scanrefer}
Dave~Zhenyu Chen, Angel~X Chang, and Matthias Nie{\ss}ner.
\newblock Scanrefer: 3d object localization in rgb-d scans using natural
  language.
\newblock In {\em European Conference on Computer Vision}, pages 202--221.
  Springer, 2020.

\bibitem{chen2021scan2cap}
Zhenyu Chen, Ali Gholami, Matthias Nie{\ss}ner, and Angel~X Chang.
\newblock Scan2cap: Context-aware dense captioning in rgb-d scans.
\newblock In {\em Proceedings of the IEEE/CVF Conference on Computer Vision and
  Pattern Recognition}, pages 3193--3203, 2021.

\bibitem{dai2017scannet}
Angela Dai, Angel~X Chang, Manolis Savva, Maciej Halber, Thomas Funkhouser, and
  Matthias Nie{\ss}ner.
\newblock Scannet: Richly-annotated 3d reconstructions of indoor scenes.
\newblock In {\em Proceedings of the IEEE conference on computer vision and
  pattern recognition}, pages 5828--5839, 2017.

\bibitem{datta2019align2ground}
Samyak Datta, Karan Sikka, Anirban Roy, Karuna Ahuja, Devi Parikh, and Ajay
  Divakaran.
\newblock Align2ground: Weakly supervised phrase grounding guided by
  image-caption alignment.
\newblock In {\em Proceedings of the IEEE/CVF International Conference on
  Computer Vision}, pages 2601--2610, 2019.

\bibitem{deng2021transvg}
Jiajun Deng, Zhengyuan Yang, Tianlang Chen, Wengang Zhou, and Houqiang Li.
\newblock Transvg: End-to-end visual grounding with transformers.
\newblock In {\em Proceedings of the IEEE/CVF International Conference on
  Computer Vision}, pages 1769--1779, 2021.

\bibitem{dionisio20133d}
John David~N Dionisio, William G~Burns III, and Richard Gilbert.
\newblock 3d virtual worlds and the metaverse: Current status and future
  possibilities.
\newblock {\em ACM Computing Surveys (CSUR)}, 45(3):1--38, 2013.

\bibitem{faghri2017vse++}
Fartash Faghri, David~J Fleet, Jamie~Ryan Kiros, and Sanja Fidler.
\newblock Vse++: Improving visual-semantic embeddings with hard negatives.
\newblock {\em arXiv preprint arXiv:1707.05612}, 2017.

\bibitem{feng2021cityflow}
Qi Feng, Vitaly Ablavsky, and Stan Sclaroff.
\newblock Cityflow-nl: Tracking and retrieval of vehicles at city scale by
  natural language descriptions.
\newblock {\em arXiv preprint arXiv:2101.04741}, 2021.

\bibitem{gupta2020contrastive}
Tanmay Gupta, Arash Vahdat, Gal Chechik, Xiaodong Yang, Jan Kautz, and Derek
  Hoiem.
\newblock Contrastive learning for weakly supervised phrase grounding.
\newblock In {\em European Conference on Computer Vision}, pages 752--768.
  Springer, 2020.

\bibitem{he2021transrefer3d}
Dailan He, Yusheng Zhao, Junyu Luo, Tianrui Hui, Shaofei Huang, Aixi Zhang, and
  Si Liu.
\newblock Transrefer3d: Entity-and-relation aware transformer for fine-grained
  3d visual grounding.
\newblock In {\em Proceedings of the 29th ACM International Conference on
  Multimedia}, pages 2344--2352, 2021.

\bibitem{he2020momentum}
Kaiming He, Haoqi Fan, Yuxin Wu, Saining Xie, and Ross Girshick.
\newblock Momentum contrast for unsupervised visual representation learning.
\newblock In {\em Proceedings of the IEEE/CVF conference on computer vision and
  pattern recognition}, pages 9729--9738, 2020.

\bibitem{hinton2015distilling}
Geoffrey Hinton, Oriol Vinyals, Jeff Dean, et~al.
\newblock Distilling the knowledge in a neural network.
\newblock {\em arXiv preprint arXiv:1503.02531}, 2(7), 2015.

\bibitem{huang2021text}
Pin-Hao Huang, Han-Hung Lee, Hwann-Tzong Chen, and Tyng-Luh Liu.
\newblock Text-guided graph neural networks for referring 3d instance
  segmentation.
\newblock In {\em Proceedings of the AAAI Conference on Artificial
  Intelligence}, volume~35, pages 1610--1618, 2021.

\bibitem{huang2022multi}
Shijia Huang, Yilun Chen, Jiaya Jia, and Liwei Wang.
\newblock Multi-view transformer for 3d visual grounding.
\newblock In {\em Proceedings of the IEEE/CVF Conference on Computer Vision and
  Pattern Recognition}, pages 15524--15533, 2022.

\bibitem{ilse2018attention}
Maximilian Ilse, Jakub Tomczak, and Max Welling.
\newblock Attention-based deep multiple instance learning.
\newblock In {\em International conference on machine learning}, pages
  2127--2136. PMLR, 2018.

\bibitem{kamath2021mdetr}
Aishwarya Kamath, Mannat Singh, Yann LeCun, Gabriel Synnaeve, Ishan Misra, and
  Nicolas Carion.
\newblock Mdetr-modulated detection for end-to-end multi-modal understanding.
\newblock In {\em Proceedings of the IEEE/CVF International Conference on
  Computer Vision}, pages 1780--1790, 2021.

\bibitem{kazemzadeh2014referitgame}
Sahar Kazemzadeh, Vicente Ordonez, Mark Matten, and Tamara Berg.
\newblock Referitgame: Referring to objects in photographs of natural scenes.
\newblock In {\em Proceedings of the 2014 conference on empirical methods in
  natural language processing (EMNLP)}, pages 787--798, 2014.

\bibitem{liu2019adaptive}
Xuejing Liu, Liang Li, Shuhui Wang, Zheng-Jun Zha, Dechao Meng, and Qingming
  Huang.
\newblock Adaptive reconstruction network for weakly supervised referring
  expression grounding.
\newblock In {\em Proceedings of the IEEE/CVF International Conference on
  Computer Vision}, pages 2611--2620, 2019.

\bibitem{liu2021group}
Ze Liu, Zheng Zhang, Yue Cao, Han Hu, and Xin Tong.
\newblock Group-free 3d object detection via transformers.
\newblock In {\em Proceedings of the IEEE/CVF International Conference on
  Computer Vision}, pages 2949--2958, 2021.

\bibitem{loshchilov2018fixing}
Ilya Loshchilov and Frank Hutter.
\newblock Fixing weight decay regularization in adam.
\newblock 2018.

\bibitem{maron1997framework}
Oded Maron and Tom{\'a}s Lozano-P{\'e}rez.
\newblock A framework for multiple-instance learning.
\newblock {\em Advances in neural information processing systems}, 10, 1997.

\bibitem{mittal2020attngrounder}
Vivek Mittal.
\newblock Attngrounder: Talking to cars with attention.
\newblock In {\em European Conference on Computer Vision}, pages 62--73.
  Springer, 2020.

\bibitem{mystakidis2022metaverse}
Stylianos Mystakidis.
\newblock Metaverse.
\newblock {\em Encyclopedia}, 2(1):486--497, 2022.

\bibitem{neubeck2006efficient}
Alexander Neubeck and Luc Van~Gool.
\newblock Efficient non-maximum suppression.
\newblock In {\em 18th International Conference on Pattern Recognition
  (ICPR'06)}, volume~3, pages 850--855. IEEE, 2006.

\bibitem{plummer2015flickr30k}
Bryan~A Plummer, Liwei Wang, Chris~M Cervantes, Juan~C Caicedo, Julia
  Hockenmaier, and Svetlana Lazebnik.
\newblock Flickr30k entities: Collecting region-to-phrase correspondences for
  richer image-to-sentence models.
\newblock In {\em Proceedings of the IEEE international conference on computer
  vision}, pages 2641--2649, 2015.

\bibitem{qi2019deep}
Charles~R Qi, Or Litany, Kaiming He, and Leonidas~J Guibas.
\newblock Deep hough voting for 3d object detection in point clouds.
\newblock In {\em proceedings of the IEEE/CVF International Conference on
  Computer Vision}, pages 9277--9286, 2019.

\bibitem{qi2017pointnet++}
Charles~Ruizhongtai Qi, Li Yi, Hao Su, and Leonidas~J Guibas.
\newblock Pointnet++: Deep hierarchical feature learning on point sets in a
  metric space.
\newblock {\em Advances in neural information processing systems}, 30, 2017.

\bibitem{rohrbach2016grounding}
Anna Rohrbach, Marcus Rohrbach, Ronghang Hu, Trevor Darrell, and Bernt Schiele.
\newblock Grounding of textual phrases in images by reconstruction.
\newblock In {\em European Conference on Computer Vision}, pages 817--834.
  Springer, 2016.

\bibitem{song2015sun}
Shuran Song, Samuel~P Lichtenberg, and Jianxiong Xiao.
\newblock Sun rgb-d: A rgb-d scene understanding benchmark suite.
\newblock In {\em Proceedings of the IEEE conference on computer vision and
  pattern recognition}, pages 567--576, 2015.

\bibitem{wang2021improving}
Liwei Wang, Jing Huang, Yin Li, Kun Xu, Zhengyuan Yang, and Dong Yu.
\newblock Improving weakly supervised visual grounding by contrastive knowledge
  distillation.
\newblock In {\em Proceedings of the IEEE/CVF Conference on Computer Vision and
  Pattern Recognition}, pages 14090--14100, 2021.

\bibitem{wang2019reinforced}
Xin Wang, Qiuyuan Huang, Asli Celikyilmaz, Jianfeng Gao, Dinghan Shen,
  Yuan-Fang Wang, William~Yang Wang, and Lei Zhang.
\newblock Reinforced cross-modal matching and self-supervised imitation
  learning for vision-language navigation.
\newblock In {\em Proceedings of the IEEE/CVF Conference on Computer Vision and
  Pattern Recognition}, pages 6629--6638, 2019.

\bibitem{xia2018gibson}
Fei Xia, Amir~R Zamir, Zhiyang He, Alexander Sax, Jitendra Malik, and Silvio
  Savarese.
\newblock Gibson env: Real-world perception for embodied agents.
\newblock In {\em Proceedings of the IEEE conference on computer vision and
  pattern recognition}, pages 9068--9079, 2018.

\bibitem{yang2020improving}
Zhengyuan Yang, Tianlang Chen, Liwei Wang, and Jiebo Luo.
\newblock Improving one-stage visual grounding by recursive sub-query
  construction.
\newblock In {\em European Conference on Computer Vision}, pages 387--404.
  Springer, 2020.

\bibitem{yang2021sat}
Zhengyuan Yang, Songyang Zhang, Liwei Wang, and Jiebo Luo.
\newblock Sat: 2d semantics assisted training for 3d visual grounding.
\newblock In {\em Proceedings of the IEEE/CVF International Conference on
  Computer Vision}, pages 1856--1866, 2021.

\bibitem{yu2018mattnet}
Licheng Yu, Zhe Lin, Xiaohui Shen, Jimei Yang, Xin Lu, Mohit Bansal, and
  Tamara~L Berg.
\newblock Mattnet: Modular attention network for referring expression
  comprehension.
\newblock In {\em Proceedings of the IEEE Conference on Computer Vision and
  Pattern Recognition}, pages 1307--1315, 2018.

\bibitem{yuan2021instancerefer}
Zhihao Yuan, Xu Yan, Yinghong Liao, Ruimao Zhang, Sheng Wang, Zhen Li, and
  Shuguang Cui.
\newblock Instancerefer: Cooperative holistic understanding for visual
  grounding on point clouds through instance multi-level contextual referring.
\newblock In {\em Proceedings of the IEEE/CVF International Conference on
  Computer Vision}, pages 1791--1800, 2021.

\bibitem{zhao20213dvg}
Lichen Zhao, Daigang Cai, Lu Sheng, and Dong Xu.
\newblock 3dvg-transformer: Relation modeling for visual grounding on point
  clouds.
\newblock In {\em Proceedings of the IEEE/CVF International Conference on
  Computer Vision}, pages 2928--2937, 2021.

\end{thebibliography}
}

\newpage
\appendix

\section{Qualitative Study}  
\label{sec:appendix_qualitative}

\begin{figure*}[tb]
    \centering
    \includegraphics[width=1\textwidth]{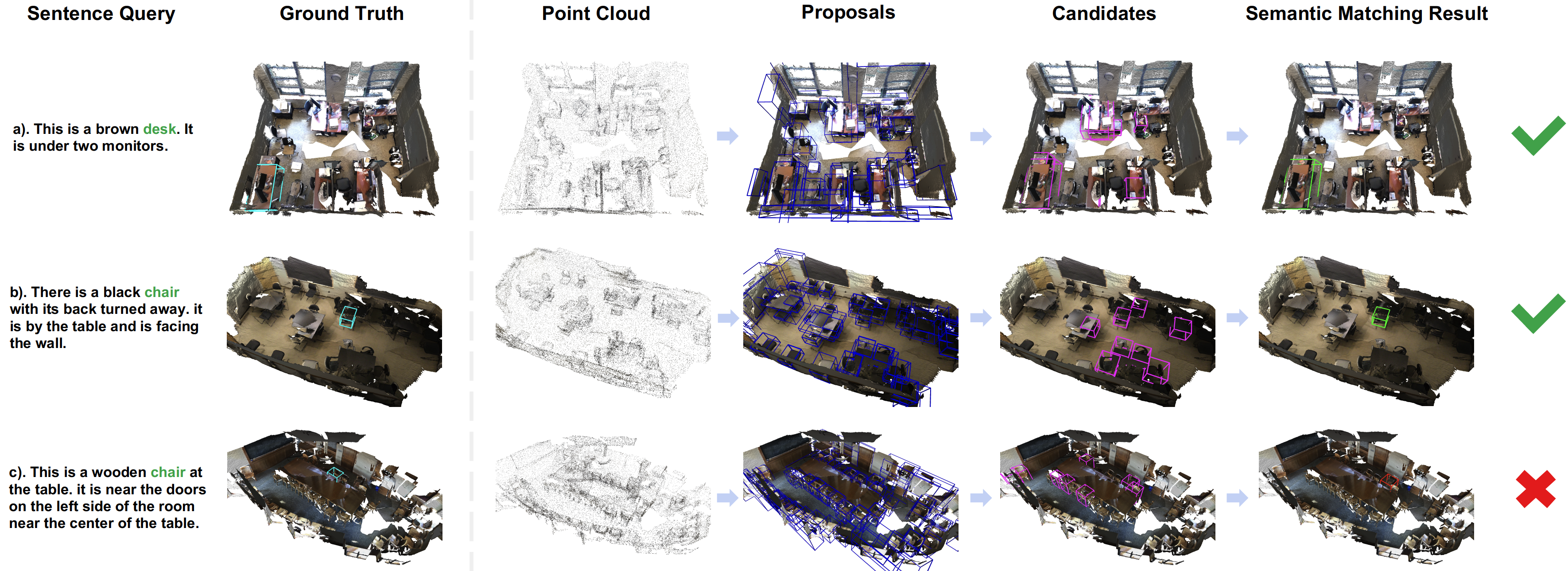}
    \caption{Visualization of the coarse-to-fine semantic matching process. With knowledge distillation, we can directly get matching result from proposals during inference (skipping the time-consuming coarse-to-fine matching).}
    \label{fig:quali_2}
\end{figure*}
    
\subsection{Coarse-to-fine Visualization}
Figure~\ref{fig:quali_2} shows the visualization results of the coarse-to-fine semantic matching pipeline. Given the input point cloud, we first extract $M_\mathrm{p}=256$ proposals by the 3D object detector. (When doing visualization, we employ the Non-Maximum Supperssion\cite{neubeck2006efficient} algorithm on these proposals to filter out some duplicate or overlapping boxes.) Then we conduct the coarse-grained candidate selection among these proposals to get $K=8$ candidates. These candidates do the fine-grained semantic matching with the sentence query one by one to finally produce the best-match result.

In all three cases, the coarse-grained module effectively selects 8 candidates related to the target object (mostly belonging to the target category). For example, in case (b), the sentence asks for a black chair near a table and facing the wall. Firstly, the module filters out objects (\textit{tables}, \textit{doors}) that do not belong to the target category (\textit{chair}). Secondly, with the assistance of the feature similarity matrix, the module's selection is also consistent with the sentence description (``by the table'' and ``facing the wall'') to some degree.

After coarse-grained selection, the reconstruction-based semantic matching process aims to differentiate these candidates in a fine-grained way. We get the correct answer in both cases (a) and (b), but we fail in case (c). Actually, in case (c), the candidates we get do not contain the target object. Considering that there are 32 different chairs in this scene, it's very likely that the 8 candidates miss the target object since we are not expecting the coarse-grained part to have a deep understanding of the objects in the same category (\textit{chair}). Even in such a condition, the fine-grained part still gets the best-match \textit{chair} (``near the doors'', ``on the left side'', and ``near the center'') among the 8 candidates, which demonstrates our model's strong ability in semantic matching.

\section{Implementation Details}

\subsection{Model Setting}
In this section, we give a detailed description of our model setting. We consider the whole two-stage 3DVG pipeline as off-the-shelf modules. In our practice, we use the pre-trained GroupFree\cite{liu2021group} model as the 3D object detector, which contains a PointNet++\cite{qi2017pointnet++} backbone, 6 layers of transformer decoder and the proposal predict head. The proposal predict head outputs the bounding boxes and the class predictions of $M_\mathrm{p}=256$ object proposals. Then we employ the same attribute encoder, textual encoder and predict module as the supervised 3DJCG\cite{cai20223djcg} model. The attribute encoder aggregates the attribute features (27-dimensional box center and corner coordinates and the 128-dimensional multi-view RGB features) and the initial object features (produced by the object detector) with 2-layer multi-head self-attention modules. The input sentence query is firstly encoded by a GloVe module, and then input to a GRU cell, which produces the text feature. The predict module is simply a 1-layer multi-head cross-attention module between the text feature (Key \& Value) and the object features (Query). For the reconstruct module, We mask the input sentence with a random ratio $p=0.3$. We use NLTK to parse the sentences, and the verbs and nouns are treated as important words. The core of the reconstruct module is a 3-layer transformer decoder. The input point number $N_\mathrm{p}$, the proposal number $M_\mathrm{p}$, and the candidate number $K$ are set to 50000, 256 and 8, respectively. The dimension of all hidden layers is 288. When calculating the candidates' rewards, we find that instead of reducing from one to zero linearly in steps of $1/(K-1)$, applying a square operation over them is better for increasing the discrimination between the optimal and suboptimal candidates.

\subsection{Training and Inference} 
We follow the weakly supervised setting where none of the object-sentence and bounding box annotations are used during training. We follow \cite{zhao20213dvg} to use 8 sentence queries for each scene to accelerate the training process. It takes 20 epochs to train our framework with a batch size of 12 ($i.e.$ there are 96 sentence queries from 12 point clouds in each batch). For the stability of the reconstruction module, we start its training at the second epoch and ignore the highest $K/2$ reconstruction losses after the third epoch. The learning rate is set to 1e-3 with cosine annealing strategy. We employ AdamW optimizer\cite{loshchilov2018fixing} with the weight decay of 5e-4. The hyper-parameters $\lambda_1$, $\lambda_2$ and $\lambda_3$ are set to 2, 2 and 1, respectively.

\end{document}